\documentclass[runningheads,a4paper]{llncs}

\usepackage{amssymb}
\usepackage{graphicx}
\usepackage{epsfig}
\usepackage{amsmath}
\usepackage{amsfonts}
\usepackage{arydshln}
\usepackage{url}
\usepackage{textpos}
\usepackage{hyperref}

\urldef{\mailsa}\path|{alfred.hofmann, ursula.barth, ingrid.haas, frank.holzwarth,|
\urldef{\mailsb}\path|anna.kramer, leonie.kunz, christine.reiss, nicole.sator,|
\urldef{\mailsc}\path|erika.siebert-cole, peter.strasser, lncs}@springer.com|    
\newcommand{\keywords}[1]{\par\addvspace\baselineskip
\noindent\keywordname\enspace\ignorespaces#1}
\begin{document}

\mainmatter  

\title{Faster K-Means Cluster Estimation}

\titlerunning{Faster K-Means Cluster Estimation}

%
%
\author{Siddhesh Khandelwal, Amit Awekar}

%

\institute{Indian Institute of Technology Guwahati\\\email{\small siddhesh166@gmail.com, awekar@iitg.ernet.in}}

%
%

\toctitle{Faster K-Means Cluster Estimation}
\tocauthor{Author Name}
 
\maketitle

\begin{abstract}
There has been considerable work on improving popular clustering algorithm `K-means' in terms of mean squared error (MSE) and speed, both. However, most of the k-means variants tend to compute distance of each data point to each cluster centroid for every iteration. We propose a fast heuristic to overcome this bottleneck with only marginal increase in MSE. We observe that across all iterations of K-means, a data point changes its membership only among a small subset of clusters. Our heuristic predicts such clusters for each data point by looking at nearby clusters after the first iteration of k-means. We augment well known variants of k-means with our heuristic to demonstrate effectiveness of our heuristic. For various synthetic and real-world datasets, our heuristic achieves speed-up of up-to 3 times when compared to efficient variants of k-means.

\keywords{K-means, Clustering, Heuristic}
\end{abstract}

\section{Introduction}
K-means is a popular clustering technique that is used in diverse fields such as humanities, bio-informatics, and astronomy. Given a dataset $D$ with $n$ data points in $\mathbb{R}^d$ space, K-means partitions $D$ into $k$ clusters with the objective to minimize the mean squared error (MSE). MSE is defined as the sum of the squared distance of each point from its corresponding centroid. The K-means problem is NP-hard. Polynomial time heuristics are commonly applied to obtain a local minimum.

One such popular heuristic is the Lloyd's algorithm\cite{lloyd1982least} that selects certain initial centroids (also referred as seeds) at random from the dataset. Each data point is assigned to the cluster corresponding to the closest centroid. Each centroid is then recomputed as mean of the points assigned to that cluster. This procedure is repeated until convergence. Each iteration involves $n*k$ distance computations. Our contribution is to reduce this cost to $n*k'$, ($k'<< k$) by generating candidate cluster list (CCL) of size $k'$ for each data point. The heuristic is based on the observation that across all iterations of K-means, a data point changes its membership only among a small subset of clusters. Our heuristic considers only a subset of nearby cluster as candidates for deciding membership for a data point. This heuristic has advantage of speeding up K-means clustering with marginal increase in MSE. We show effectiveness of our heuristic by extensive experimentation using various synthetic and real-world datasets.


\section{Our Work: Candidate cluster list for each data point} \label{ourWork}
Our main contribution is in defining a heuristic that can be used as augmentation to current variants of k-means for faster cluster estimation. Let algorithm $V$ be a variant of k-means and algorithm $V'$ be the same variant augmented with our heuristic. Let $T$ be the time required for $V$ to converge to MSE value of $E$. Similarly, $T'$ is the time required for $V'$ to converge to MSE value of $E'$. We should satisfy following two conditions when we compare $V$ with $V'$:
\begin{itemize}
    \item Condition 1: $T'$ is lower than $T$, and
    \item Condition 2: $E'$ is either lower or only marginally higher than $E$.
\end{itemize}
In short, these conditions state that a K-means variant augmented with our heuristic should converge faster without significant increase in final MSE.

Major bottleneck of K-means clustering is the computation of data point to cluster centroid distance in each iteration of K-means. For a dataset with $n$ data points and $k$ clusters, each iteration of K-means performs $n*k$ such distance computations. To overcome this bottleneck, we maintain a CCL of size $k'$ for each data point. We assume that $k'$ is significantly smaller than $k$. We discuss the effect of various choices for the size of CCL in Section \ref{experiments}. We build CCL based on top $k'$ nearest clusters to the data point after first iteration of K-means. Now each iteration of K-means will perform only $n*k'$ distance computations.

Consider a data point $p_{1}$ and cluster centroids represented as $c_{1}, c_{2} ... , c_{k}$. Initially all centroids are chosen randomly or using one of the seed selection algorithms mentioned in Section \ref{relatedWork}. Let us assume that $k'$= 4, and $k'<<k$. After first iteration of K-means $c_{8}, c_{5}, c_{6}, \text{ and } c_{1}$ are the top four closest centroids to $p_{1}$ in the increasing order of distance. This is the candidate cluster list for $p_{1}$. If we run K-means for second iteration, $p_{1}$ will compute distance to all $k$ centroids. After second iteration, top four closest centroid list might change in two ways:
\begin{enumerate}
    \item Members of the list do not change but only ranking changes among the members. For example, top four closest centroid list for $p_{1}$ might change to $c_{1}, c_{6}, c_{8}$, and $c_{5}$ in the increasing order of distance.
    \item Some of the centroids in the previous list are replaced with other centroids which were not in the list. For example, top four closest list for p1 might change to $c_{5}, c_{2}, c_{9}, \text{ and } c_{8} $ in the increasing order of distance
\end{enumerate}

For many synthetic and real world datasets we observe that the later case rarely happens. That is, the set of top few closest centroids for a data point remains almost unchanged even though order among them might change. Therefore, CCL is a good enough estimate for the closest cluster when K-means converges \cite{additionalResults}. For each data point, our heuristic involves computation overhead of $O(k.log(k))$ for creating CCL and memory overhead of $O(k')$ to maintain CCL. For a sample dataset consisting 100,000 points in 54 dimensions and the value of $k = 100$ and $k'= 40$, this overhead is approximately 30MB.


\section{Related Work} \label{relatedWork}

In last three decades, there has been significant work on improving Lloyd's algorithm \cite{lloyd1982least} both in terms of reducing MSE and running time. The follow up work on Lloyd's algorithm can be broadly divided into three categories: Better seed selection\cite{arthur2007k,likas2003global}, Selecting ideal value for number of clusters\cite{pham2005selection}, and Bounds on data point to cluster centroid distance\cite{elkan2003using,pelleg1999accelerating,kanungo2002efficient}. Arthur et. al.\cite{arthur2007k} provided a better method for seed selection based on a probability distribution over closest cluster centroid distances for each data point. Likas et. al.\cite{likas2003global} proposed the Global k-means method for selecting one seed at a time to reduce final mean squared error. Pham et. al.\cite{pham2005selection} designed a novel function to evaluate goodness of clustering for various potential values of number of clusters. Elkan\cite{elkan2003using} use triangle inequality to avoid redundant computations of distance between data points and cluster centroids. Pelleg and Moore\cite{pelleg1999accelerating} and Kanungo et al.\cite{kanungo2002efficient} proposed similar algorithms that use k-d trees. Both these algorithms construct a k-d tree over the dataset to be clustered. Though these approaches have shown good results, k-d trees perform poorly for datasets in higher dimensions.

Seed selection based K-means variants differ from Lloyd's algorithm only in the method of seed selection. Our heuristic can be directly used in such algorithms. K-means variants that find appropriate number of clusters in data, evaluate the goodness of clustering for various potential values of number of clusters. Such algorithms can use our heuristic while performing clustering for each potential value of $k$. K-means variants in third category compute exact distances only to few centroids for each data point. However, they have to compute bounds on distances to rest of the centroids for each data point. Our heuristic can help such K-means variants to further reduce distance and bound calculations.

\begin{table*}[t]
\centering
\caption{Datasets used in experiment}
\label{datasets}
\begin{tabular}{|l|c|c|l|}
\hline
\textbf{Name}      & \textbf{Cardinality} & \textbf{Dimensionality} & \textbf{Description}                                             \\ \hline
Birch     & 100000      & 2              & 10 by 10 grid of Gaussian clusters \\ \hline
Covtype   & 150000      & 55             & remote soil cover measurements            \\ \hline
Mnist     & 60000       & 784            & original NIST handwritten digit training data           \\ \hline
KDDCup    & 95412       & 481            & KDD Cup 1998 data                                       \\ \hline
Synthetic & 100000      & 100            & Uniform random dataset                              \\ \hline
\end{tabular}
\end{table*}

\begin{table*}[t]
\centering
\caption{Effect of varying $k'$ on HT performance. The value of $k$ = 100. \newline\hspace{\textwidth} RND = Random initialization ; KPP = Initialization using Kmeans++\cite{arthur2007k}}
\label{varyingkprime}
\begin{tabular}{ll:cc:cc:cc:cc:cc}
\hline
\textbf{}  & \textbf{}            & \multicolumn{2}{c:}{\textbf{$k'$ = 20}} & \multicolumn{2}{c:}{\textbf{$k'$ = 30}} & \multicolumn{2}{c:}{\textbf{$k'$ = 40}} & \multicolumn{2}{c:}{\textbf{$k'$ = 50}} & \multicolumn{2}{c}{\textbf{$k'$ = 60}} \\
           &                      & RND                & KPP               & RND                & KPP               & RND                & KPP               & RND                & KPP               & RND                & KPP               \\ \hline
Birch      & PIM(\%) & -0.11              & 0                 & 0.04               & 0                 & 0                  & 0                 & 0                  & 0                 & 0                  & 0                 \\
\hdashline & Speedup              & 3.05               & 3.14              & 2.48               & 2.26              & 2.01               & 1.93              & 1.68               & 1.67              & 1.41               & 1.31              \\ \hline
Covtype    & PIM(\%) & 0.21               & 0.03              & 0.02               & 0                 & 0                  & 0                 & 0                  & 0                 & 0                  & 0                 \\
\hdashline & Speedup              & 2.32               & 2.02              & 1.81               & 1.82              & 1.61               & 1.63              & 1.55               & 1.38              & 1.42               & 1.20              \\ \hline
Mnist      & PIM(\%) & 1.30               & 1.36              & 0.60               & 0.71              & 0.36               & 0.36              & 0.30               & 0.18              & 0.23               & 0.09              \\
\hdashline & Speedup              & 1.89               & 1.47              & 1.60               & 1.44              & 1.42               & 1.26              & 1.38               & 1.19              & 1.37               & 1.15              \\ \hline
KDDCup     & PIM(\%) & 0.81               & 0.70              & 0.11               & 0.15              & 0.08               & 0.02              & -0.18              & -0.01             & 0                  & 0                 \\
\hdashline & Speedup              & 1.44               & 1.60              & 1.33               & 1.15              & 1.42               & 1.02              & 0.88               & 0.99              & 1.18               & 1.02              \\ \hline
Synthetic  & PIM(\%) & 0.19               & 0.15              & 0.11               & 0.08              & 0.06               & 0.04              & 0.03               & 0.01              & 0.01               & 0.01              \\
\hdashline & Speedup              & 2.90               & 2.45              & 2.28               & 1.97              & 1.87               & 1.71              & 1.51               & 1.35              & 1.36               & 1.17              \\ \hline
\end{tabular}
\end{table*}

\section{Experimental Results} \label{experiments}

Our heuristic can be augmented to multiple variants of K-means mentioned in Section \ref{relatedWork}. When augmented to Lloyd's algorithm, our heuristic provides a speedup of around 30 times with the error within 0.2\% of that of Lloyd's algorithm. However to show the effectiveness of our heuristic, we present results of augmenting it to faster variants of K-means such as K-means with triangle inequality (KMT)\cite{elkan2003using}. Due to lack of space, we present results of augmenting our heuristic with only this variant. Augmenting KMT with our heuristic is referred as algorithm HT. Code and datasets used for our experiments are available for download \cite{additionalResults}.

During each iteration of KMT, a data point computes distance to the centroid of its current cluster. KMT uses triangle inequality to compute efficient lower bounds on distances to all other centroids. A data point will compute exact distance to any other centroid only when the lower bound on such distance is smaller than the distance to the centroid of its current cluster. During each iteration of HT, a data point will also compute distance to the centroid of its current cluster. However, HT will compute lower bounds on distances to centroids only in its CCL. A data point will compute exact distance to any other centroid in the candidate cluster list only when the lower bound on such distance is smaller than the distance to the centroid of its current cluster.

Experimental results are presented on five datasets, four of which were used by Elkan et. al.\cite{elkan2003using} to demonstrate the effectiveness of KMT and one is a synthetically generated dataset by us. These datasets vary in dimensionality from 2 to 784, indicating applicability of our heuristic for low as well as high dimensional data (please refer to Table \ref{datasets}). Our evaluation metrics are chosen based on two conditions mentioned in Section \ref{ourWork}: Speedup to satisfy Condition 1 and Percentage Increase in MSE (PIM) to satisfy Condition 2. Speedup is calculated as $T/T'$. PIM is calculated as $(100 * (E' -E))/E$. We tried two different methods for initial seed selection: random \cite{lloyd1982least}and K-means++ \cite{arthur2007k}. Both seed selection methods gave similar trends in results. To ensure fair comparison, the same initial seeds are used for both KMT and HT. For some experiments, HT achieves smaller MSE than KMT ($E'\leq E$). This happens because our heuristic jumps the local minima by not computing distance to every cluster centroid. Only in such cases, HT requires more iterations to converge and runs slower than KMT.

\underline{\textbf{Effect of Varying $k'$}}: Please refer to Table \ref{varyingkprime}. The value of the total number of clusters $k$ is set to 100 for all datasets. Running time and MSE of KMT is independent of value of $k'$. Speed up of HT over KMT increases with reduction in value of $k'$. This is expected as for small value of $k'$, HT can avoid many redundant distance computations using small CCL. Speed up of HT over KMT is not same as the ratio $k/k'$. Reason for reduced speed up is that KMT also avoids some distance computations using its own filtering criteria of triangle inequality. Our heuristic achieves ideal speed of $k/k'$ when compared against basic K-means algorithm \cite{additionalResults}. $E'$ increases with reduction in value of $k'$. However, $E'$ is only marginally higher than $E$ as PIM value never exceeds 1.5.

\underline{\textbf{Effect of Varying $k$}}: Please refer to Table \ref{varyingk}. Here, we report results for value of $k'$ set to $0.4 * k$. With increasing value of $k$, HT achieves better speed up over KMT and difference between MSE of HT and MSE of KMT reduces. With increasing value of $k$, most of the centroid to data point distance calculations become redundant as data-point is assigned only to the closest centroid. In such scenario, our heuristic avoids distance computations with reduced PIM. This shows that our heuristic can be used for datasets having only few as well as large number of clusters.

\begin{table}[t]
\centering
\caption{Effect of varying $k$ on HT performance. The value of $k' = 0.4 * k$. \newline\hspace{\textwidth} RND = Random initialization ; KPP = Initialization using Kmeans++\cite{arthur2007k}}
\label{varyingk}
\begin{tabular}{ll:cc:cc:cc:cc}
\hline
\textbf{}  & \textbf{}            & \multicolumn{2}{c:}{\textbf{$k$ = 50}} & \multicolumn{2}{c:}{\textbf{$k$ = 100}} & \multicolumn{2}{c:}{\textbf{$k$ = 500}} & \multicolumn{2}{c}{\textbf{$k$ = 1000}} \\
           &                      & RND               & KPP               & RND                & KPP               & RND                & KPP               & RND                & KPP                \\ \hline
Birch      & PIM(\%)       & 0.31              & 0                 & 0                  & 0                 & 0                  & 0                 & 0                  & 0                  \\
\hdashline & Speedup   & 1.65              & 1.71              & 1.98               & 1.97              & 2.14               & 2.10              & 2.12               & 2.15               \\ \hline
Covtype    & PIM(\%)       & 0.01              & 0.02              & 0.26               & 0                 & 0                  & 0                 & 0                  & 0                  \\
\hdashline & Speedup   & 1.35              & 1.31              & 1.65               & 1.50              & 1.94               & 1.87              & 1.97               & 1.90               \\ \hline
Mnist      & PIM(\%)       & 0.94              & 0.87              & 0.38               & 0.52              & 0.09               & 0.23              & 0.13               & 0.07               \\
\hdashline & Speedup   & 1.10              & 1.20              & 1.23               & 1.45              & 1.28               & 1.24              & 1.29               & 1.19               \\ \hline
KDDCup     & PIM(\%)       & 0.51              & 0.99              & -0.06              & 0.15              & 0                  & 0.03              & 0                  & 0.02               \\
\hdashline & Speedup   & 1.02              & 1.38              & 0.85               & 1.18              & 1.13               & 1.33              & 1.19               & 1.37               \\ \hline
Synthetic  & PIM(\%)       & 0.09              & 0.07              & 0.05               & 0.04              & 0.03               & 0.01              & 0.01               & 0.01               \\
\hdashline & Speedup   & 2.03              & 1.63              & 1.76               & 1.56              & 1.75               & 1.45              & 1.56               & 1.51               \\ \hline
\end{tabular}
\end{table}

\underline{\textbf{Effect of Seeding}}: Please refer to Table \ref{varyingkprime} and Table \ref{varyingk}. For each value of $k'$ in Table \ref{varyingkprime} and $k$ in Table \ref{varyingk}, we used two different initial seedings - random (RND) and Kmeans++ \cite{arthur2007k}. If we compare the results, we observe that better seeding (KMeans++) generally gives better results in terms of PIM. Randomly selected seeds are not necessarily well distributed across the dataset. In such cases, successive iterations of K-means causes significant changes in cluster centroids. Improved seeding methods such as KMeans++ ensure that the initial centroids are spread out more uniformly. Thus centroids shift is less significant in successive iterations. In such scenario, CCL computed after first iteration is a better estimate for final cluster membership. Thus our heuristic is expected to perform better with newer variants of K-means that provide improved seeding.

\underline{\textbf{Effect of Cluster Well-Separateness}}: We also performed experiments on synthetic datasets in two dimensions. These datasets were generated using a mixture of Gaussians. The Gaussian centers are placed at equal angles on a circle of radius $r$ ($angle = \frac{2\pi}{k}$), and each center is assigned equal number of points ($\frac{n}{k}$). The experiment was done on synthetic datasets of 100000 points generated using the method described above with variance set to 0.25. The value of $k$ is set to 100 and the value of $k'$ is set to 40. We generated nine datasets by varying the radius from zero to forty in steps of five units. We ran KMT and HT over these nine datasets to check how our heuristic performs with change in well separateness of clusters. We observed that when clusters are close, both the algorithms converge quickly as initial seeds happen to be close to actual cluster centroids. With higher radius, initial seeds might be far off from the actual cluster centroids and KMT takes longer to converge. However, HT performs significantly better for higher values of radius as HT can quickly discard far away clusters. HT achieves a speedup of around 2.31 for higher radius values. For all experiments over these synthetic datasets, we observed that PIM value never exceeds 0.01\cite{additionalResults}. This indicates that our heuristic remains relevant even with variation in degree of separation among the clusters.

\section{Conclusion}
We presented a heuristic to attack the bottleneck of redundant distance computations in K-means. Our heuristic limits distance computations for each data point to CCL. Our heuristic can be augmented with diverse variants of K-means to converge faster without any significant increase in MSE. With extensive experiments on real-world and synthetic datasets, we showed that our heuristic performs well with variations in dataset dimensionality, CCL size, number of clusters, and degree of separation  among clusters. This work can be further improved by making the CCL dynamic to achieve better speed up while reducing the PIM value.

\bibliographystyle{abbrv}
\bibliography{sigprop} 

\begin{thebibliography}{1}

\bibitem{additionalResults}
The code and dataset for the experiments can be found at:\\
  \url{https://github.com/siddheshk/Faster-Kmeans}.

\bibitem{arthur2007k}
D.~Arthur and S.~Vassilvitskii.
\newblock k-means++: The advantages of careful seeding.
\newblock In {\em ACM-SIAM symposium on Discrete algorithms}, pages 1027--1035,
  2007.

\bibitem{elkan2003using}
C.~Elkan.
\newblock Using the triangle inequality to accelerate k-means.
\newblock In {\em International Conference om Machine Learning}, pages
  147--153, 2003.

\bibitem{kanungo2002efficient}
T.~Kanungo, D.~M. Mount, N.~S. Netanyahu, C.~D. Piatko, R.~Silverman, and A.~Y.
  Wu.
\newblock An efficient k-means clustering algorithm: Analysis and
  implementation.
\newblock {\em Pattern Analysis and Machine Intelligence, IEEE Trans. on},
  24(7):881--892, 2002.

\bibitem{likas2003global}
A.~Likas, N.~Vlassis, and J.~J. Verbeek.
\newblock The global k-means clustering algorithm.
\newblock {\em Pattern recognition}, 36(2):451--461, 2003.

\bibitem{lloyd1982least}
S.~P. Lloyd.
\newblock Least squares quantization in pcm.
\newblock {\em Information Theory, IEEE Trans. on}, 28(2):129--137, 1982.

\bibitem{pelleg1999accelerating}
D.~Pelleg and A.~Moore.
\newblock Accelerating exact k-means algorithms with geometric reasoning.
\newblock In {\em ACM SIGKDD}, pages 277--281. ACM, 1999.

\bibitem{pham2005selection}
D.~T. Pham, S.~S. Dimov, and C.~Nguyen.
\newblock Selection of k in k-means clustering.
\newblock {\em Journal of Mechanical Engineering Science}, 219(1):103--119,
  2005.

\end{thebibliography}








\end{document}